\documentclass{article}
\usepackage{ijcai24}

\usepackage{times}
\usepackage{soul}
\usepackage{url}
\usepackage[hidelinks]{hyperref}
\usepackage[utf8]{inputenc}
\usepackage[small]{caption}
\usepackage{graphicx}
\usepackage{amsmath}
\usepackage{amsthm}
\usepackage{booktabs}
\usepackage{algorithm}
\usepackage{algorithmic}
\usepackage[switch]{lineno}
\usepackage{amssymb} 
\usepackage{subcaption}
\usepackage{forest}

\newtheorem{theorem}{Theorem}
\newtheorem{proposition}{Proposition}
\newcommand{\equal}[1]{{\hypersetup{linkcolor=black}\thanks{#1}}}

\title{Why Attention Graphs Are All We Need: Pioneering Hierarchical Classification of Hematologic Cell Populations with LeukoGraph}

\author{
Fatemeh Nassajian Mojarrad$^1$\equal{These authors contributed equally}
\and
Lorenzo Bini$^1$\footnotemark[1]\and
Thomas Matthes$^2$\And
Stéphane Marchand-Maillet$^1$\\
\affiliations
$^1$Department of Computer Science, University of Geneva, Switzerland\\
$^2$Hematology Service, Department of Oncology and Clinical Pathology Service, Department of Diagnostics, Geneva University Hospital, Switzerland\\
\emails
\{fatemeh.nassajian, lorenzo.bini, stephane.marchand-maillet\}@unige.ch,
thomas.matthes@hcuge.ch
}
\begin{document}

\maketitle

\begin{abstract}
In the complex landscape of hematologic samples such as peripheral blood or bone marrow, cell classification, delineating diverse populations into a hierarchical structure, presents profound challenges. This study presents LeukoGraph, a recently developed framework designed explicitly for this purpose employing graph attention networks (GATs) to navigate hierarchical classification (HC) complexities. Notably, LeukoGraph stands as a pioneering effort, marking the application of graph neural networks (GNNs) for hierarchical inference on graphs, accommodating up to one million nodes and  millions of edges, all derived from flow cytometry data. LeukoGraph intricately addresses a classification paradigm where for example four different cell populations undergo flat categorization, while a fifth diverges into two distinct child branches, exemplifying the nuanced hierarchical structure inherent in complex datasets. 
The technique is more general than this example.
A hallmark achievement of LeukoGraph is its F-score of 98\%, significantly outclassing prevailing state-of-the-art methodologies. Crucially, LeukoGraph's prowess extends beyond theoretical innovation, showcasing remarkable precision in predicting both flat and hierarchical cell types across flow cytometry datasets from 30 distinct patients. This precision is further underscored by LeukoGraph's ability to maintain a correct label ratio, despite the inherent challenges posed by hierarchical classifications.
\end{abstract}

\section{Introduction}
Flow cytometry is a technology that provides rapid multi-parametric analysis of single cells in solution. The expression of specific cell surface and intracellular molecules is thereby detected by antibodies coupled to fluorochromes. Exposing labelled cells to a laser leads to the emission of fluorescence which is then collected by a detector.  By analyzing millions of cells in a short time information can thereby be collected about the presence or absence of up to fifty different cellular molecules per cell.  
This technique is employed in medicine to assess the cellular composition of complex body fluids like blood or bone marrow and to facilitate the diagnosis of hematologic diseases such as leukemia. 
We have generated a dataset obtained by the analysis of bone marrow samples from 30 patients with information obtained from 20212 to 133820 cells/sample about the presence or absence of ten different surface molecules. The pattern of expression of these molecules allows to distinguish in each sample seven different cell populations:  T lymphocytes, B lymphocytes, monocytes, mast cells, hematopoietic stem and progenitor cells (HSPC), and two subpopulations of HSPCs, myeloid and lymphoid HSPCs (see Fig.  \ref{fig:tree}).


The primary research focus in machine learning has predominantly been centered on developing models for conventional classification problems, wherein an object is assigned to a single class from a set of disjoint classes. However, a distinct subset of tasks involves scenarios where classes are not disjoint  but rather organized hierarchically, giving rise to HC. In HC, objects are linked to a specific superclass along with its corresponding subclasses. Depending on the nature of the task, this association may encompass all subclasses or only a designated subset. The hierarchical structure formalizing the interrelation among classes can manifest as either a tree or a directed acyclic graph (DAG) \cite{Silla2011}.
Hierarchical classification problems manifest across a broad spectrum, spanning from musical genre classification 
\cite{Ariyaratne2012,Iloga2018}
to identification of COVID-19
in chest X-ray images \cite{Pereira2020},
taxonomic classification of
viral sequences in metagenomic data
\cite{Shang2021} and
text categorization \cite{Javed2021,Ma2022}.
In this context, each prediction must adhere to the hierarchical constraint, ensuring coherence with the organizational structure.
The hierarchy constraint dictates that a data point assigned to a particular class must also be associated with all its ancestors in the hierarchical structure.

In this paper we present a LeukoGraph, an approach tailored for HC problems.
This method utilizes a network $\mathcal{H}$ designed for the underlying  classification problem, effectively harnessing hierarchy information to generate predictions that adhere to the hierarchy constraint, ultimately enhancing overall performance \cite{Giunchiglia2020}.
In order to integrate the LeukoGraph, we proceed in two steps. First,
in order to guarantee  coherent predictions through construction, we build a constraint layer  on top of $\mathcal{H}$. Then we train LeukoGraph with a loss function to determine when to leverage predictions from the lower classes in the hierarchy to make predictions on the upper ones.

Graph-based models have the capability to leverage both the global and local characteristics inherent in networks relevant  to cell biology.
 GNNs have been recently adapted in various
tasks, e.g. link prediction \cite{Zhang2018}, graph classification \cite{Duvenaud2015,Lee2019},
and node classification \cite{Kipf2017,Velickovic2017,Yang2016}.
Several GNNs have demonstrated effectiveness by achieving state-of-the-art performance across diverse graph datasets, including but not limited to citation networks \cite{Velickovic2017}, recommender systems \cite{Kipf2017,Yang2016} and protein-protein interaction networks \cite{Velickovic2017}.
GNNs leverage the underlying graph structure to perform convolution directly on graphs, either by transmitting node features to neighbors \cite{Yang2016} or by executing convolution in the spectral domain by using the eigenbasis of the graph Laplacian operator \cite{Kipf2017}.
 \cite{Vaswani2017}  introduced the self-attention mechanism.
From there, we can handle model with variable-sized inputs and focus on the most important part of the inputs  by using self-attention.  
Self-attention mechanism has been introduced to graphs as GAT by \cite{Velickovic2017} and further improved by subsequent works 
\cite{Bottou2005,Choi2017,Yun2019}. In this paper we build $\mathcal{H}$ as a GAT
to perform HC.

The outline of the manuscript is as follows: in Section \ref{sec:Data} we  introduce the HC problem via the depiction of the data we face  in our work. In Section \ref{sec:Methodology} we describe the LeukoGraph and the architecture of the network $\mathcal{H}$
for the underlying classification problem.
In Section \ref{sec:Experiments} we test the validity and effectiveness of the proposed strategy, namely,
we will test LeukoGraph on flow cytometry data.
Finally, Section \ref{sec:Conclusion} provides some
conclusions and future perspectives.

\section{Problem Description}
\label{sec:Data}
Flow cytometry data is tabular data where each row corresponds to a single cell and every column to a marker or feature (cell surface molecule -- Table \ref{tab:1}). \begin{table}[H]
\scriptsize
    \centering
    \begin{tabular}{ccc}
        \hline
       Marker \# & Marker  & Explanation  \\
        \hline
       0&  FS INT & Forward Scatter (FSC) - Cell’s size \\
       1& SS INT & Side Scatter (SSC) - Cell’s granularity\\
       2&  CD14-FITC & Cluster of Differentiation 14 - Antigen\\
       3& CD19-PE & Cluster of Differentiation 19 - Antigen\\
       4& CD13-ECD &Cluster of Differentiation 13 - Antigen \\
      5&  CD33-PC5.5 & Cluster of Differentiation 33 - Antigen\\
      6&  CD34-PC7 & Cluster of Differentiation 34 - Antigen\\
      7&  CD117-APC &Cluster of Differentiation 117 - Antigen \\
       8& CD7-APC700 &Cluster of Differentiation 7 - Antigen \\
      9&  CD16-APC750 &Cluster of Differentiation 16 - Antigen \\
      10&  HLA-PB &Human Leukocyte Antigen \\
       11& CD45-KO & Cluster of Differentiation 45 - Antigen\\
        \hline
\end{tabular}
    \caption{Flow cytometry data markers.}
    \label{tab:1}
\end{table}

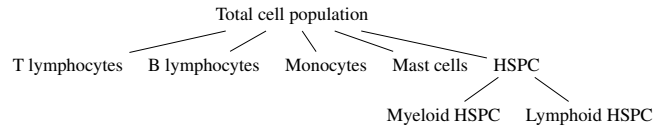
\begin{figure}[h]
\centering
\centering  
\scalebox{0.7}{
\begin{forest}
  [Total cell population[T lymphocytes][B lymphocytes][Monocytes][Mast cells][HSPC[Myeloid HSPC][Lymphoid HSPC], triangle]]
\end{forest}}
\caption[Depiction of the  HC model.]{Depiction of the HC model.}
\label{fig:tree}
\end{figure}

According to the expression pattern of all the markers on a single cell, the cell can be classified into a specific category (cell population):
\begin{enumerate}
    \item T lymphocytes: a type of white blood cell that plays a crucial role in the adaptive immune system.
    \item B lymphocytes: a type of white blood cell that plays a crucial role in the adaptive immune system.
    \item Monocytes: a type of white blood cell that plays a crucial role in the innate immune system.
    \item Mast cells: a type of immune cell that plays a crucial role in the body's response to allergies and certain infections (also part of the innate immune system).
    \item  HSPC: a population of immature cells from which the other cell populations are derived.
    \item Myeloid HSPC: a subset of HSPCs that gives rise to the myeloid lineage of blood cells, such as monocytes and mast cells. 
    \item Lymphoid HSPC: a subset of HSPCs that gives rise to the lymphoid lineage of blood cells, such as B and T lymphocytes.
\end{enumerate}
Biologic knowledge indicates that these cell populations   are further organized into a hierarchy, as shown in Fig. \ref{fig:tree}.

This setup therefore formally gives rise to the following research question: "How to leverage class hierarchy during classification?".

\section{Methodology}
\label{sec:Methodology}
Given a tabular dataset containing $n$ data samples and $m$ feature fields, denoted by $X=[\mathbf{x}_1, \cdots,\mathbf{x}_n]^T$, the $i$-th sample in the table associated with $m$ feature values $\mathbf{x}_i=[ x^{(1)}_i, \cdots,x^{(m)}_i]$ and a discrete label $y_i$,  part of a label hierarchy. 
Our learning goal is to find a mapping function $f$ for which given $\mathbf{x}_i$, returns 
the predicted label $\hat{y}_i$. In the rest of this section, we will describe  LeukoGraph  and the architecture of the
network for the underlying  classification problem.
\subsection{LeukoGraph}
We consider a HC problem with a given set $\mathcal{C}$ of $C$ classes, which are hierarchically organized as a tree \cite{Giunchiglia2020}. 
If there exists a path of non-negative length from class $A_1$ to class $A_2$ in the tree, we call $A_2$  a subclass of $A_1$. 

We assume to have a mapping $f_A: \mathbb{R}^m \to [0,1]$ for every class $A$ such that $\mathbf{x}\in \mathbb{R}^m$ is predicted to belong to $A$ whenever $f_A(\mathbf{x})$ is greater than or equal to some threshold.

When $A_2$ is a subclass of $A_1$, the model should guarantee that $f_{A_1}(\mathbf{x})\geq f_{A_2}(\mathbf{x})$, for all $\mathbf{x}\in \mathbb{R}^m$, in order to  guarantee that the hierarchy constraint is always satisfied independently from the threshold. 
Note that in this model, every class is considered a subclass of itself.

The case where $f_{A_1}(\mathbf{x})<f_{A_2}(\mathbf{x})$ for some $\mathbf{x}\in \mathbb{R}^m$ and $A_2$ is a subclass of $A_1$, is called a {\em hierarchy violation} \cite{Wehrmann2018}.

\paragraph{Overall structure} Given a set $\mathcal{C}=\{A_1,\cdots,A_C\}$ of $C$ hierarchically structured classes $A_c$, we first build a neural network composed of two modules:  
a {\em Early Module} $\mathcal{H}$ with one output for each class in $\mathcal{C}$ and an upper module, referred to as the {\em Max Constraint Module} (MCM), consisting of a single layer that takes the output of the early module as input and imposes the hierarchy constraint.

If $y_A$ is the ground truth label for class $A\in \mathcal{C}$ and the prediction value $\mathcal{H}_A \in [0,1]$ is obtained by $\mathcal{H}$ for class $A$, the MCM output for a class $A$ is given by:
\begin{equation}
\label{MCM}
    \text{MCM}_A=\max_{B\in \mathcal{S}_A} \mathcal{H}_B.
\end{equation}
where $\mathcal{S}_A$ is the set of subclasses of $A$ in $\mathcal{C}$. 
Note that since the number of operations performed by $\text{MCM}_A$ is independent from the depth of the hierarchy, for each class $A \in \mathcal{C}$, the resulting model is scalable. 

By construction, the following theorem therefore holds.
\begin{theorem}
\label{theorem1}
Let $\mathbf{x}\in \mathbb{R}^m$ be a data point.
Let $\mathcal{C}=\{A_1,\cdots, A_C\}$  be a set of hierarchically structured classes and let $\mathcal{H}$ be a early module with outputs $\mathcal{H}_{A_1},\cdots, \mathcal{H}_{A_C}$ ($\mathcal{H}_{A_c}\in[0,1]~\forall c$) given the input $\mathbf{x}$, and $\text{MCM}_{A_1},\cdots, \text{MCM}_{A_C}$ be defined as in equation \eqref{MCM}. 
Then LeukoGraph does not admit hierarchy violations.
\end{theorem}
Second, in order to  exploit the hierarchy constraint during training, LeukoGraph  is trained with {\em max constraint loss} (MCLoss), defined as follows
\begin{equation*}
   \text{MCLoss}_A=-y_A \ln (\max_{B\in \mathcal{S}_A} y_B \mathcal{H}_B ) -(1-y_A) \ln (1-\text{MCM}_A),
\end{equation*}
for each class $A\in \mathcal{C}$.
The global MCLoss is written as 
\begin{equation}
  \text{MCLoss}= \sum_{A\in \mathcal{C}}  \text{MCLoss}_A.
\end{equation}
The advantage of defining MCLoss instead of using the standard binary cross-entropy loss is that the more ancestors a class has, the more likely it is that LeukoGraph trained with the standard binary cross-entropy loss will remain stuck in spurious local optima. 
The MCLoss prevents this to happen.

We also use a {\em weighted loss function}.
The most common way to implement a weighted loss function is to assign higher weights to minority classes and lower weights to majority classes. 
These weights are often set to be inversely proportional to the frequency of classes, ensuring our conditions.

As a result, LeukoGraph has the ability of delegating the prediction for a class $A_i$ to one of its subclasses $A_j$, thanks to MCM and the MCLoss.
More formally, the following proposition holds by construction.
\begin{proposition}
Under the same assumptions as in Theorem~\ref{theorem1}, considering a class $A_i\in \mathcal{C}$ and $A_j\in \mathcal{S}_{A_i}$ with $i \neq j$, LeukoGraph delegates the prediction on $A_i$ to $A_j$ for $\mathbf{x}$, if $\text{MCM}_{A_i}=\mathcal{H}_{A_j}$ and $\mathcal{H}_{A_i}<\mathcal{H}_{A_j}$.
\end{proposition}
\paragraph{Classification layer}
We built the Early Module $\mathcal{H}$ for the classification as a Graph Attention Network (GAT), and remind of its structure below, following \cite{Velickovic2017} (section 2.1).

Given $\{\mathbf{x}_i\}_{i=1}^n$ as nodes, where $n$ is the number of nodes and $\mathbf{x}_i\in \mathbb{R}^m$ with $m$ the number of features in each node, the node features are the input of the graph attention layer.  
The output of the layer is a new set of node features $\{\mathcal{H}_i\}_{i=1}^n$, where $\mathcal{H}_i \in \mathbb{R}^{\Tilde{m}}$, for which $\Tilde{m}$ might be different from $m$.

\begin{center}
\begin{figure}[H]
\includegraphics[width=0.45\textwidth]{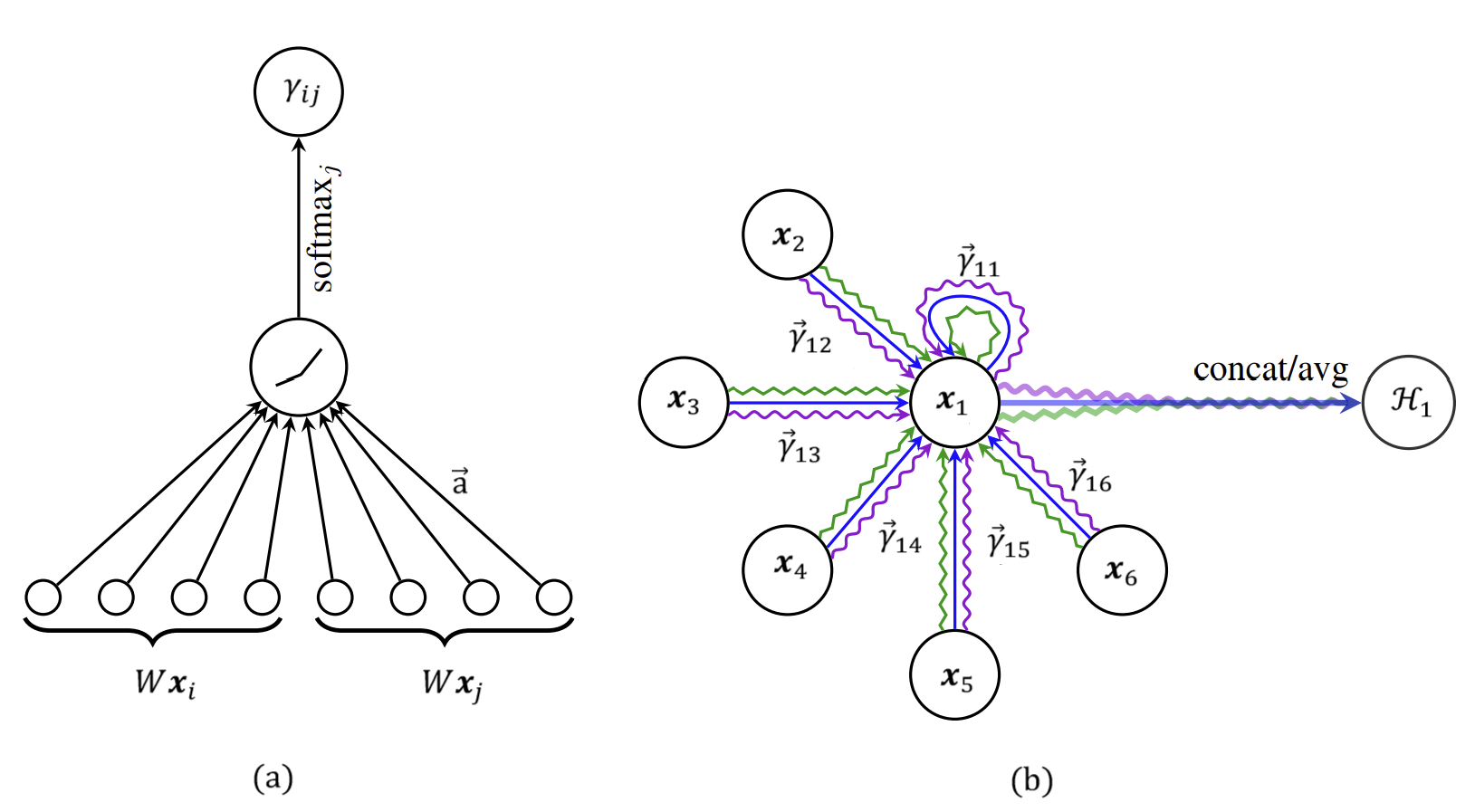}
	\caption{(a): Computing the normalized attention coefficients $\gamma_{ij}$  (b): Multi-head attention of node 1 on its neighborhood. Arrows show concatenation or averaging of attention}
 	\label{fig:network}
\end{figure}
\end{center}

The main difference between a GAT and a graph convolutional network (GCN) \cite{Kipf2017} is that it defines {\em attention coefficients} as importance scores of node $\mathbf{x}_j$'s features to that of node $\mathbf{x}_i$.
$W$ is the weight matrix used to build attention coefficients.
{\em Masked attention } normalizes {\em attention coefficients} within every graph neighborhood $\mathcal{N}(\mathbf{x}_i)$.
In our case, $\mathcal{N}(\mathbf{x}_i)$ represents the Euclidean $k$-nearest neighbours of node $\mathbf{x}_i$.
The attention mechanism is a single-layer feedforward neural network, parameterized by a weight vector $\vec{a} \in \mathbb{R}^{2\Tilde{m}}$, over which we apply the LeakyReLU non-linearity (Fig. \ref{fig:network}(a)). 

The normalized attention coefficients $\gamma_{ij}$ are then used to compute a linear combination of the  features corresponding to them, after applying a nonlinearity $\sigma$, to serve as the final output features for each node. 

We also move from single head attention to multi-head attention (with $L$ heads) by concatenating the output of all heads  at every layer $i$ (but the final layer).
The expression for the final layer replaces concatenation by averaging and can be written in our notation as
\begin{equation}
 \mathcal{H}_i= \sigma (\frac{1}{L}\sum_{l=1}^L \sum_{j \in \mathcal{N}(\mathbf{x}_i)} \gamma_{ij}^l W^l\mathbf{x}_j).   
\end{equation}
Fig. \ref{fig:network}(b) recaps the structure of the multi-head attention mechanism (adapted from \cite{Velickovic2017}). 

\section{Experiments} 
\label{sec:Experiments}
In this section, we present the experimental results that illustrate the behavior of LeukoGraph with GAT for the underlying classification problem on flow cytometry data. 
We also compare the proposed method with current state-of-the-art models.
\subsection{Flow Cytometry Data Preparation}
 The flow cytometry data introduced in Section \ref{sec:Data} is analyzed by flow cytometry experts  and each cell is manually mapped into a category based on several 2D projections of the data.
 As a result of overlap, a cell may receive two or more annotations. 
 We opted to duplicate such a cell with each copy receiving a single annotation.

 We also normalize each feature and map its $[\min,\max]$ interval to $[0,1]$.

 Our complete dataset comprises 30 patients with an average of $n=69144$ cells per patient and $m=12$ markers.
 Average class distribution ratios, across 30 patients are shown in Table~\ref{tab:ratios}.
 \begin{table}[H]
\footnotesize
    \centering
    \begin{tabular}{cc}
        \hline
        Label  & Average concentration\\
        \hline
        T lymphocytes  & 61.03\\
        B lymphocytes  & 13.18\\
        Monocytes  & 15.41\\
        Mast cells &0.21 \\
        HSPC & 10.17\\
        Myeloid HSPC &7.03 \\
        Lymphoid HSPC &2.60\\
        \hline
    \end{tabular}
    \caption{Average class distribution ratios across 30 patients.}
    \label{tab:ratios}
\end{table}
\subsection{Setup}
In our experiments, we construct the $k$-nearest neighbors graphs with $k=5$ neighbors. 
We perform a 7-fold test procedure where 1/7th of the data (4 or 5 patients out of 30) is held out as a test set and the rest (6/7th of data) as a training set. We then perform a 10-fold cross validation procedure within the training set to set the hyperparameters.
The structure of the model has the following hyperparameters: 2 layers, 64 hidden channels, and 8 attention heads.
The model is trained with an initial learning rate of $\eta=0.1$ and the dropout parameter is set to 0.4. 
The nonlinearity $\sigma$ for the first layer is set to ReLU.
\subsection{Comparaison with other methods}
The metrics are reported  using the extensions of the renowned metrics of precision, recall and F-score, but tailored to the HC setup. 
We implement the metrics of hierarchical precision ($hp$), hierarchical
 recall ($hr$) and hierarchical F-score ($hf$) defined by \cite{Kiritchenko2006}
\begin{equation*}
    hp=\frac{\sum_i |\alpha_i \cap \beta_i|}{\sum_i |\alpha_i|}, ~hr=\frac{\sum_i |\alpha_i \cap \beta_i|}{\sum_i |\beta_i|},~ hf=\frac{2\times hp\times hr}{hp+ hr}.
\end{equation*}
where $\alpha_i$ is the set consisting of the most specific classes predicted for test sample $i$ and all their ancestor classes and $\beta_i$ is the set containing the true most specific classes of test sample $i$ and all their ancestors.  

\begin{figure}[H]
\begin{center}
\includegraphics[width=0.4\textwidth]{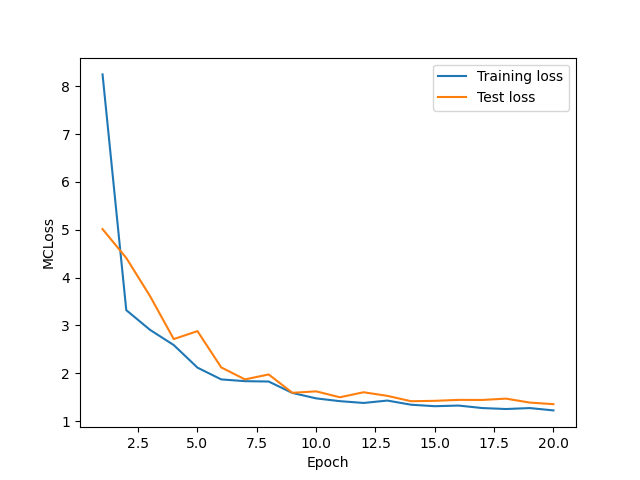}
	\caption{MCLoss for training and test
sets.
	\label{fig:train_val_loss}}
\end{center}
\end{figure}

Fig.~\ref{fig:train_val_loss} shows the plots of training loss and test loss, indicating no overfitting. We compare LeukoGraph with a deep neural network (DNN) and graph deep learning classifiers, e.g. GCN and graph neural network (GNN). In Table \ref{tab:4} the performance  is measured in terms of $hp$, $hr$ and $hf$. 
 The average ratios of predicted label across all patients is shown in Table \ref{tab:5}.
In our experiments, we  observe that the results for the LeukoGraph strategy  are much better than the DNN and other graph  deep learning classifiers.
We also present the results obtained with XGBoost, Gaussian mixture model (GMM) and random forest (RF) in Appendix \ref{appendix:2}.

\begin{table}[H]
\footnotesize
    \centering
    \begin{tabular}{ccccc}
        \hline
        Metrics  & LeukoGraph  
        &GCN & GNN & DNN \\
        \hline
        hp  &\textbf{0.983} 
        & 0.979&0.972&0.807\\
        hr  & \textbf{0.985}
        &0.980& 0.976 &0.771\\
        hf  &\textbf{0.984} 
        &0.980 & 0.974&0.788\\
        \hline
\end{tabular}
    \caption{Average metrics across all patients using different  networks.}
    \label{tab:4}
\end{table}

\begin{table}[H]
\footnotesize
    \centering
    \begin{tabular}{ccccc}
        \hline
        Label  & LeukoGraph  
        &GCN & GNN & DNN\\
        \hline
        T lymphocytes  & 99.28 
        &99.55& 99.67&\textbf{100.00}\\
        B lymphocytes  & \textbf{87.82} 
        &82.67&76.34 &-\\
        Monocytes  & 97.11 
        &\textbf{97.45} &94.87&-\\
        Mast cells & \textbf{83.01} 
        &78.50 &73.11 &-\\
        HSPC & \textbf{82.09} 
        & 75.39&77.25&-\\
        Myeloid HSPC & \textbf{89.43} 
        &86.39&87.80&-\\
        Lymphoid HSPC & \textbf{73.02}  
        & 62.67&53.35&-\\
        \hline
    \end{tabular}
    \caption{Average ratios across all patients using different  networks.}
    \label{tab:5}
\end{table}

\begin{figure*}
\begin{subfigure}{.5\textwidth}
  \centering  \includegraphics[width=.9\linewidth]{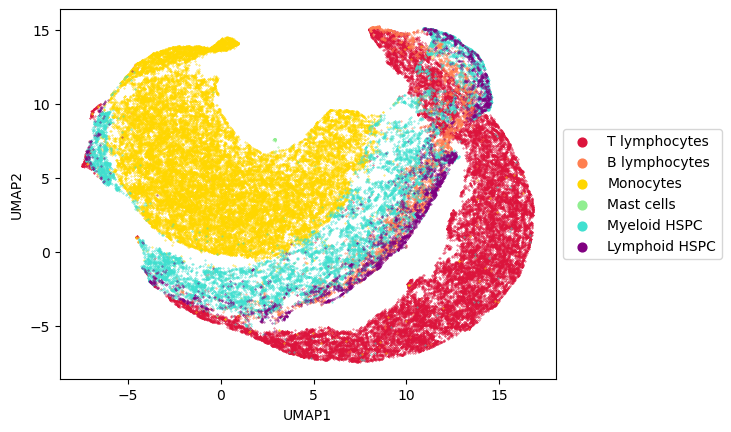}
  \caption{UMAP scatter plot}
 \label{fig:fig11_1}
\end{subfigure}%
\begin{subfigure}{.5\textwidth}
 \centering
 \includegraphics[width=.7\linewidth]{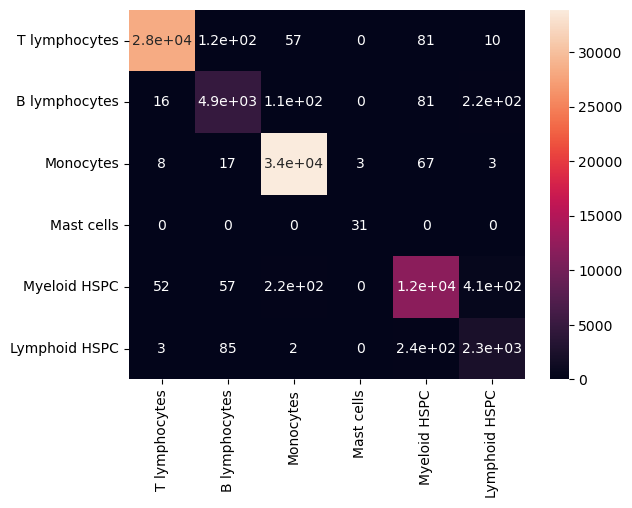}
  \caption{Confusion matrix}
  \label{fig:fig11_2}
\end{subfigure}
\caption{Results for patient 11}
\label{fig:fig11}
\end{figure*}

\begin{figure*}
\begin{subfigure}{.5\textwidth}
  \centering  \includegraphics[width=.9\linewidth]{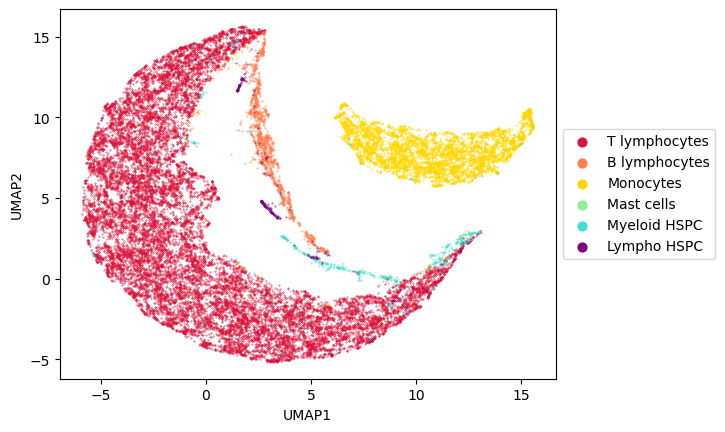}
  \caption{UMAP scatter plot}
 \label{fig:fig12_1}
\end{subfigure}%
\begin{subfigure}{.5\textwidth}
 \centering
 \includegraphics[width=.7\linewidth]{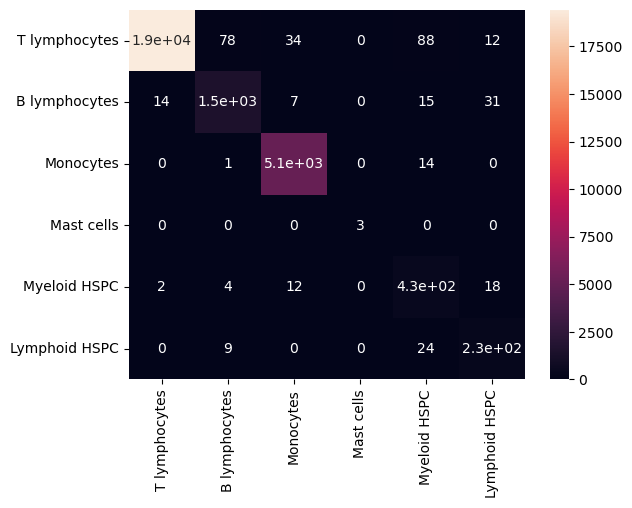}
  \caption{Confusion matrix}
  \label{fig:fig12_2}
\end{subfigure}
\caption{Results for patient 12}
\label{fig:fig12}
\end{figure*}

\begin{figure*}
\begin{subfigure}{.5\textwidth}
  \centering  \includegraphics[width=.9\linewidth]{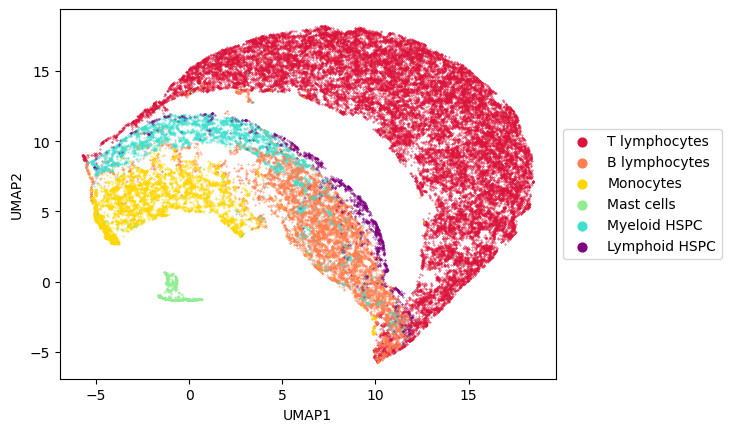}
  \caption{UMAP scatter plot}
 \label{fig:fig23_1}
\end{subfigure}%
\begin{subfigure}{.5\textwidth}
 \centering
 \includegraphics[width=.7\linewidth]{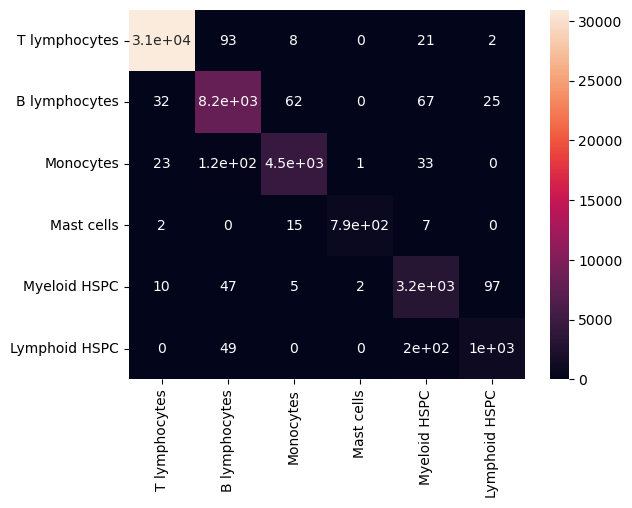}
  \caption{Confusion matrix}
  \label{fig:fig23_2}
\end{subfigure}
\caption{Results for patient 23}
\label{fig:fig23}
\end{figure*}

Figs. \ref{fig:fig11}, \ref{fig:fig12} and \ref{fig:fig23} demonstrate the performance of  LeukoGraph for three random patients. In the left panels, we show the UMAP scatter plot colored by various cell types.
In the right panels, we show the confusion matrix. 

\begin{figure}
    \centering
\includegraphics[width=0.5\textwidth]{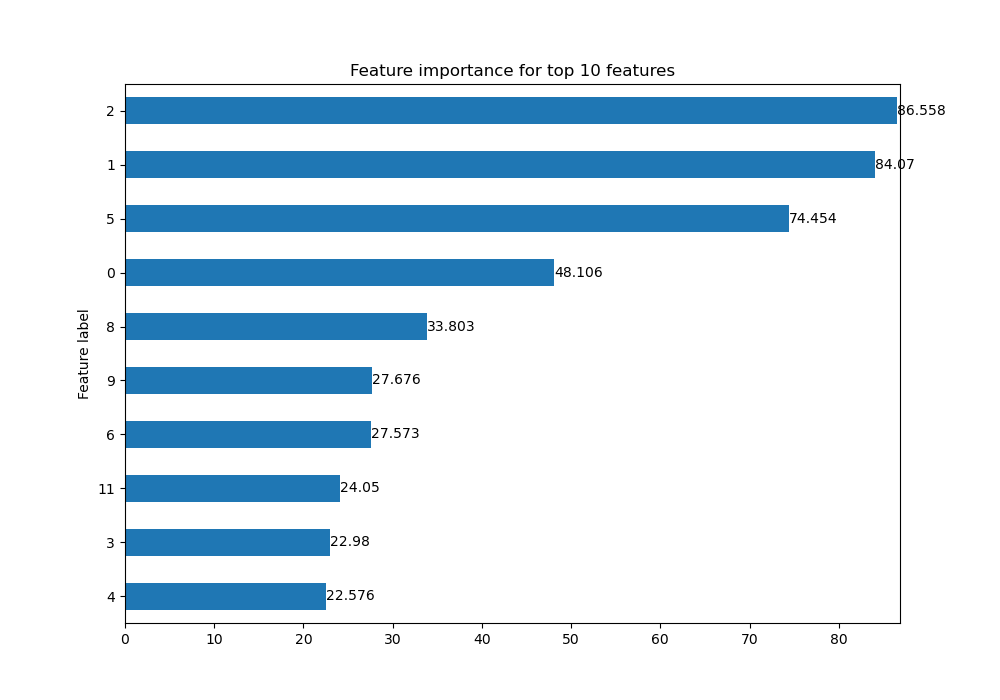}
    \caption{Feature weighting attributed by $\mathcal{H}$.}
    \label{fig:features}
\end{figure}

To finally demonstrates that our models paves a way to interpretability, Fig. \ref{fig:features} show the features importance. We discuss all of these results in the following section. 
\subsection{Discussion of Our Results}
The results presented in this paper underscore the novelty and effectiveness of our newly introduced hierarchical GNNs framework, LeukoGraph, in performing node classification tasks. This general approach leverages the inherent hierarchical structure of the data, a feature that has been largely overlooked in traditional GNNs and other deep learning classifiers, but frequently appears in real-word dataset.

An important aspect of our LeukoGraph model that warrants special attention is its ability to accurately capture and classify even small populations of certain cell classes. This is a significant achievement, as it demonstrates the recall and precision of our model. These performances of LeukoGraph are further illustrated through case studies of three random patients (Figs. \ref{fig:fig11}, \ref{fig:fig12}, and \ref{fig:fig23}). The UMAP scatter plots and confusion matrices provide a visual testament to the model’s ability to accurately classify different cell types.

Therefore, our hierarchical GNN framework, with its ability to capture the hierarchical relationships in the data, turn out to be particularly well-suited for this kind of task. By effectively leveraging the hierarchical structure, our model can focus on different levels of granularity in the data, thereby enabling it to detect small populations. 

Our method for managing hierarchical structure, is indeed a significant novelty in the field of GNNs node classification, since we have defined it to work on more general hierarchical dataset, where appears to be a need of integrating inherent hierarchical relationships from the data to the model. Moreover, this particular structure allows LeukoGraph to model inner relationships between different cell types during the classification process, considering not only the individual characteristics of each cell type, but also their positions and relationships within the overall hierarchy.

This is particularly important in medical applications, where certain cell types may be present in relatively small numbers but are of critical importance. However, we want to stress and to highlight that the previous illustrated one is a very general hierarchical framework, suitable to multi-class classification problem showing similar structure to Fig. \ref{fig:tree}.

Finally, an important aspect of LeukoGraph is its interpretability, as demonstrated in Fig. \ref{fig:features}. This feature importance analysis not only provides insights into the decision-making process of our model but also paves the way for potential improvements and adaptations of our model for other applications.  As we can see, the features that are given the most importance (at least above 40\%) are 2,1,5 and 0, therefore we broke down the choice of the model for each one. In order of importance, we had:
\begin{itemize}
    \item CD14-FITC: This feature reflects the level of CD14 expression, a receptor found on the cell surface that interacts with lipopolysaccharide (LPS), a component of bacterial cell walls. Predominantly found on monocytes, macrophages, and activated granulocytes, CD14 plays a pivotal role in mediating the immune response to bacterial infections. The high weightage given to this feature suggests that our model is adept at distinguishing between cells involved in innate immunity and those that aren’t, thereby potentially detecting the presence of bacterial infection in the sample.
    \item Side Scatter (SSC)-Cell’s granularity: This feature measures the level of light scatter at a 90-degree angle to the laser beam, reflecting the internal complexity or granularity of the cell. This could include the presence of granules, nuclei, or other organelles. The high weightage of this feature implies that our model can differentiate between cells of varying complexity, such as lymphocytes, monocytes, and granulocytes, or identify cells with abnormal granularity, such as blast cells or malignant cells.
    \item CD33-PC5.5: This feature indicates the expression level of CD33, a cell surface receptor of the sialic acid-binding immunoglobulin-like lectin (Siglec) family. CD33 is expressed by myeloid cells like monocytes, macrophages, granulocytes, and mast cells, and it modulates the immune response by inhibiting the activation of these cells. The high weightage of this feature suggests that our model can distinguish between myeloid and non-myeloid cells, or detect the expression of CD33 as a marker for certain types of leukemia, such as acute myeloid leukemia (AML) or chronic myelomonocytic leukemia (CMML).
    \item Forward scatter (FSC)-Cell’s size: This feature measures the level of light scatter along the path of the laser beam, which is proportional to the diameter or surface area of the cell. This can be used to discriminate cells by size. The importance placed on this feature suggests that our model can differentiate between cells of different sizes, such as small lymphocytes and large monocytes.
\end{itemize}

In conclusion, the introduction of our novel hierarchical GNN framework, LeukoGraph, marks a significant advancement in the field of node classification. By effectively leveraging the hierarchical structure of the data, our model achieves superior performance while ensuring robustness and interpretability. We believe that our work will inspire future research in this direction and contribute to the advancement of graph deep learning classifiers.

\section{Conclusion and Future Work}
\label{sec:Conclusion}
In this paper, we propose a  model,
 called LeukoGraph, for HC problems where the network $\mathcal{H}$ designed for the underlying classification problem is a GAT.
 The advantage of considering this model is that
 we can leverage the hierarchical side-information to learn when to delegate the prediction on a superclass to
one of its subclasses, 
generate predictions that are inherently coherent, and  outperform current
state-of-the-art models on flow cytometry data.

To adeptly navigate the inherent imbalances across diverse cellular populations, LeukoGraph incorporates a weighted loss structure. This design ensures equitable representation and fosters robust performance, even in the face of hierarchical complexities and imbalances. 

In general, LeukoGraph redefines HC paradigms within cellular analysis, highlighting the potential of GNNs. Through this avant-garde framework, we set a new benchmark for precision, and scalability in cell classification, charting a promising trajectory for future advancements in the domain.

There is considerable potential for artificial intelligence (AI), specifically LeukoGraph, to augment clinical practice.
The presented model is very efficient and computationally fast.
It typically takes approximately 20 to 25 minutes to manually analyze flow cytometry data obtained from a patient. In contrast, LeukoGraph accomplishes the same task in just 1-minute inference time.

In the future, we would like to enlarge the dataset from our patients to include a broader range of hematologic diseases and diverse cellular populations.
Our goal is to build a robust benchmark \cite{flowcyt2024} that encapsulates the intricacies of cellular analysis comprehensively. Additionally, we will concentrate our efforts towards exploring the potential of semi-supervised and self-supervised learning methodologies. These approaches is very efficient in terms of time  (fewer labels needed) and cost, providing a practical path for scalable advancements.

\appendix
\section{Results for Flat Classification}
\label{appendix:2}
We consider the   flat approach, where a
single multi-class classifier is trained to predict only the leaf nodes in hierarchical
data (see Fig. \ref{fig:tree}), completely ignoring previous levels of the hierarchy. 
Hence, the datasets have been  categorized  into seven distinct classes: 
T lymphocytes, B lymphocytes, monocytes,  mast cells, myeloid HSPC,  lymphoid HSPC and HSPC\textbackslash myeloid HSPC\textbackslash lymphoid HSPC.
We use the sklearn package for XGBoost, GMM and RF.
In Table \ref{tab:6}, the performance is listed  in 
terms of  precision, recall, and F1 score.
The predicted 
label across patients and cell types is shown in Table \ref{tab:7}.
Our  experiments demonstrate clearly
that using the LeukoGraph leads to a better performance.
\begin{table}[H]
\footnotesize
    \centering
    \begin{tabular}{cccc}
        \hline
        Metrics  & XGBoost &GMM & RF \\
        \hline
        Precision  &0.856 & 0.189 & 0.879 \\
        Recall  &0.846 &0.189 & 0.889 \\
        F1 Score  &0.856 &0.189 & 0.882  \\
        \hline
\end{tabular}
    \caption{Average metrics across all patients using different models.}
    \label{tab:6}
\end{table}

\begin{table}[H]
\footnotesize
    \centering
    \begin{tabular}{cccc}
        \hline
        Label  &XGBoost& GMM &RF \\
        \hline
        T lymphocytes  & 94.89 & 66.92& 81.90 \\
        B lymphocytes  & 87.67 & 95.43& 87.91 \\
        Monocytes  & 93.29&- & 88.22 \\
        Mast cells & 2.70 & 57.48& - \\
        Myeloid HSPC & 40.87 &53.80 & 23.64 \\
        Lymphoid HSPC & 39.59 &- & 3.14 \\
         HSPC\textbackslash myeloid HSPC\\  \textbackslash lymphoid HSPC & 27.85 &0.27 & - \\
        \hline
    \end{tabular}
    \caption{Average ratios across all patients using different models.}
    \label{tab:7}
\end{table}

\section*{Acknowledgments}
This work is partly funded by the Swiss National Science Foundation under grant number 207509 "Structural Intrinsic Dimensionality".


\bibliographystyle{named}
\bibliography{ijcai24}

\end{document}